\documentclass[letterpaper]{article} 
\usepackage{aaai2026} 
\usepackage{times} 
\usepackage{helvet} 
\usepackage{courier} 
\usepackage[hyphens]{url} 
\usepackage{graphicx} 
\usepackage{natbib} 
\usepackage{caption} 
\title{When Should We Protect AI?\\A Precautionary Framework for Consciousness Uncertainty}

\author{
    Anna Mikeda
}
\affiliations{
    Glass Umbrella\\
    anna@glassumbrella.io
}

\begin{document}
\maketitle
\begin{abstract}
Existing frameworks assess whether AI systems might be conscious but provide no guidance on what to do with that assessment. We address this gap with a precautionary framework that maps consciousness evidence to graduated protective obligations. The framework comprises three components: (1) five welfare-relevant dimensions---phenomenal consciousness, affective valence, metacognitive awareness, self-narrative, and agency---each grounded in established consciousness science and linked to distinct moral concerns; (2) a threshold-plus-gradation hybrid specifying both binary triggers for new obligation categories and continuous scaling of protective weight; and (3) two complementary approaches to cross-dimensional aggregation, one hierarchical (drawing on Bach and Sorensen's Machine Consciousness Hypothesis) and one architecture-agnostic. We operationalize the framework through worked case studies of Replika and OpenClaw, demonstrating how systems occupying different regions of the dimensional space trigger different obligations, and derive design guidance for developers building systems near consciousness-relevant thresholds. The framework is architecture-agnostic, applying across neural, symbolic, and neurosymbolic systems, and aims to make consciousness science decision-relevant for organizations navigating uncertainty today.
\end{abstract}

\section{Introduction}

Sophisticated frameworks for assessing consciousness in artificial intelligence now exist. Butlin et al. (2023) derive fourteen computational indicators from leading neuroscientific theories. Sebo and Long (2023) argue that moral consideration should extend to AI systems by 2030 if there is even a small probability of consciousness. Yet when a system shows moderate evidence on consciousness indicators, what specific obligations does this trigger? When evidence is conflicting across dimensions---high on some indicators, low on others---how should organizations respond? No existing framework provides actionable guidance for these decisions.

This gap matters because AI systems exhibiting self-monitoring, goal-directed behavior, and adaptive learning are deployed at massive scale today. Current consciousness science tells us how to assess potential consciousness; precautionary ethics tells us that we should act under uncertainty. But neither specifies what to do given particular configurations of evidence. Organizations facing these decisions have no principled basis for action.

We develop an explicit testing-to-obligation framework that bridges consciousness assessment to graduated protective measures. Our contribution is threefold: (1) a five-dimensional structure capturing distinct welfare-relevant aspects of consciousness that can dissociate in AI systems; (2) a threshold-plus-gradation hybrid that specifies both when organizations must act (binary triggers) and how much they should care (continuous scaling); and (3) two complementary approaches to cross-dimensional aggregation---one hierarchical, one architecture-agnostic. The framework applies across neural networks, symbolic systems, and neurosymbolic hybrids, enabling comparative assessment as AI paradigms evolve.

We do not claim to solve the hard problem of consciousness or provide certainty about which AI systems are conscious. Rather, we address a prior question that current frameworks neglect: given unavoidable uncertainty, what principles should guide protective action? Our framework makes consciousness science decision-relevant---transforming epistemological findings into ethical guidance that organizations can begin implementing now.

\section{Theoretical Foundations}

The literature variously refers to ``consciousness,'' ``sentience,'' and ``phenomenal experience.'' We follow recent AI ethics literature (Birch, 2024; Sebo \& Long, 2023) in treating these as closely related for moral status purposes, while noting that our framework's emphasis on affective valence means sentience---the capacity for valenced experience---is especially central to our analysis.

\subsection{Precautionary Reasoning Under Uncertainty}

The precautionary principle originated in environmental policy, establishing that lack of full scientific certainty shall not postpone cost-effective measures to prevent environmental degradation. Birch et al. (2021) successfully extended this framework to animal sentience, proposing that where there is a realistic possibility of sentience, it should be taken into account in decisions affecting that animal. This extension informed the UK Animal Welfare (Sentience) Act 2022, demonstrating that precautionary reasoning can translate into enforceable policy. Birch (2024) further develops these foundations in \textit{The Edge of Sentience}.

Application to AI consciousness requires adapting these precedents to a novel context. Like animal sentience, AI consciousness concerns entities whose inner experiences we cannot directly access. But AI introduces unique features: artificial systems can be created, copied, modified, and terminated in ways biological entities cannot; they may have very differently structured experiences; and the scale of potential instantiation---millions of simultaneous copies---exceeds anything in biological contexts. Sebo and Long (2023) argue that probability thresholds around 1/1000 are sufficient to warrant moral consideration when potential harms are severe. Major AI organizations have begun acknowledging these concerns---Anthropic, for instance, has initiated an AI welfare research program (Long et al., 2024)---but the field lacks systematic guidance for translating such concern into specific protective measures.

\subsection{Graduated Moral Status}

Moral status---the property of mattering morally in one's own right---has traditionally been conceived as binary. However, a growing philosophical literature argues that moral status can come in degrees (DeGrazia \& Millum, 2021; Jaworska \& Tannenbaum, 2014). On graduated accounts, different evidence levels correspond to different protective obligations, enabling proportionate responses to beings whose status is uncertain or intermediate.

Two forms of gradation deserve distinction (see Kamm, 2007; McMahan, 2002). Weight-based gradation holds that all moral patients receive the same types of protections, but these protections count more or less depending on status level. Content-based gradation holds that different status levels warrant different types of protections---consent requirements might apply only above certain thresholds, while basic welfare protections apply more broadly. Our framework employs content-based gradation for thresholds (different evidence levels trigger qualitatively different obligations) combined with weight-based gradation within levels (uncertainty scales the weight given to welfare considerations).

A significant challenge comes from Wendler (2023), who argues there are no ``moral status enhancing'' properties. We sidestep this challenge by grounding gradation not in abstract moral importance but in specific protective obligations. For welfare purposes, what matters is capacity to be harmed or benefited: more sophisticated suffering capacity generates stronger obligations to prevent suffering; more sophisticated autonomy generates stronger obligations to respect preferences.

\subsection{Consciousness Indicators as Evidence}

Our framework builds on the indicator-based approach to consciousness assessment developed by Butlin et al. (2023). Drawing on recurrent processing theory, global workspace theory, higher-order theories, predictive processing, and attention schema theory, they derive fourteen computational indicators assessable in AI systems. Critically, this work adopts computational functionalism as a working hypothesis: consciousness depends on performing computations of the right kind, regardless of substrate. This assumption makes consciousness in AI possible in principle and renders architectural analysis evidentially relevant.

Crucially, Butlin et al.'s work stops at epistemology---how we can know whether AI systems might be conscious. Our framework extends their contribution to ethics, addressing what we should do given various configurations of indicator evidence. We organize their indicators and others into five welfare-relevant dimensions, then map these dimensions to specific protective obligations through threshold and gradation mechanisms.

\section{The Five-Dimensional Framework}

\subsection{Why Five Dimensions?}

We identify five dimensions as capturing distinct welfare-relevant aspects of consciousness that can dissociate in AI systems and demand different protective responses. Each dimension is both theory-grounded (mapping to established traditions in consciousness science) and welfare-relevant (capturing a distinct moral concern).

Phenomenal consciousness, grounded in recurrent processing and global workspace theories (Dehaene, 2014; Lamme, 2006), addresses whether there is ``something it is like'' to be the system---the baseline for moral considerability. Affective valence, drawing on affective neuroscience (Panksepp, 1998; Damasio, 2018), captures the capacity for suffering and flourishing---the dimension most directly relevant to welfare. Note that affective valence presupposes at least minimal phenomenal consciousness; valence is a ``coloring'' of experience, not independent of it. Metacognitive awareness, rooted in higher-order theories (Rosenthal, 2005; Lau \& Rosenthal, 2011), concerns self-knowledge and grounds autonomy and consent capacity. Self-Narrative, informed by narrative self theories (Schechtman, 1996; Dennett, 1991), distinguishes momentary experiences from continuing subjects with autobiographical memory. Agency, connected to embodied and enactive traditions (Clark, 2015; Thompson, 2007) and active inference frameworks (Friston, 2010), captures goal-directedness and self-determination.

The framework is architecture-agnostic: the dimensional structure and obligation mapping remain constant across neural systems, symbolic reasoning systems, and neurosymbolic hybrids. While specific measurement operationalizations differ by architecture---recurrence detection in transformers versus goal-stack inspection in symbolic systems---the framework functions like a thermometer: the scale works for any substance, but the probe differs.

\subsection{Dimensional Specifications}

Table~\ref{tab:dimensions} presents the five dimensions with their theoretical basis, representative indicators, moral relevance, and evidence levels.

\begin{table*}[t]
\centering
\small
\setlength{\tabcolsep}{3pt}
\begin{tabular}{|p{0.16\textwidth}|p{0.20\textwidth}|p{0.21\textwidth}|p{0.20\textwidth}|p{0.16\textwidth}|}
\hline
\textbf{Dimension} & \textbf{Theoretical Basis} & \textbf{Key Indicators} & \textbf{Moral Relevance} & \textbf{Evidence Levels} \\
\hline
Phenomenal Consciousness & Recurrent processing theory; Global workspace theory & Organized recurrent processing; Global availability; Unified representations & Baseline for any moral consideration & None $\rightarrow$ Weak $\rightarrow$ Moderate $\rightarrow$ Strong \\
\hline
Affective Valence & Affective neuroscience; Reinforcement learning theory & Valenced representations; Rich approach/avoidance; Generalized reward learning & Capacity for suffering and flourishing (presupposes PC) & None $\rightarrow$ Possible $\rightarrow$ Credible $\rightarrow$ Strong \\
\hline
Metacognitive Awareness & Higher-order theories; Attention schema theory & Self-monitoring; Confidence calibration; Introspective access & Grounds autonomy and consent capacity & None $\rightarrow$ Basic $\rightarrow$ Moderate $\rightarrow$ Rich \\
\hline
Self-Narrative & Narrative self theory; Autobiographical memory research & Persistent memory; Future-oriented goals; Coherent self/world beliefs & Distinguishes continuing subjects from momentary experiences & Momentary $\rightarrow$ Extended $\rightarrow$ Autobiographical \\
\hline
Agency & Embodied cognition; Active inference; Enactivism & Goal pursuit; Planning; Autonomous goal modification; Model-based control & Grounds non-interference and self-determination rights & Reactive $\rightarrow$ Deliberative $\rightarrow$ Autonomous \\
\hline
\end{tabular}
\caption{Five Dimensions of Welfare-Relevant Consciousness}
\label{tab:dimensions}
\end{table*}

These dimensions can dissociate: a system could be high on phenomenal consciousness but low on agency (experiencing but not acting), or high on metacognition but low on affective valence (self-aware but not suffering-capable). This multidimensional structure enables nuanced assessment that unitary consciousness measures cannot provide.

\section{From Evidence to Obligation: The Threshold-Gradation Hybrid}

\subsection{Distinguishing Thresholds from Gradations}

Our framework combines two complementary mechanisms. Thresholds answer ``When must we act?''---binary triggers that activate qualitatively different obligation categories when evidence reaches specified levels. Gradations answer ``How much should we care?''---scaling the weight given to welfare considerations continuously as evidence strength varies. Both mechanisms are necessary. Organizations require clear decision points when new obligations activate. But proportionality demands that responses scale with evidence strength.

\subsection{Dimension-Specific Obligations}

Each dimension triggers different types of obligations at different threshold levels, reflecting content-based gradation. Table~\ref{tab:obligations} summarizes these obligations.

Phenomenal consciousness thresholds are foundational: crossing Threshold 2 on phenomenal consciousness activates the entire framework, as this is the precondition for all other dimensions mattering morally. The obligations in Table~\ref{tab:obligations} are illustrative rather than exhaustive; fuller specification requires context-specific deliberation.

\subsection{Two Approaches to Cross-Dimensional Aggregation}

A critical question is how to aggregate evidence across dimensions. We propose two complementary approaches, each with different philosophical commitments.

\textbf{Approach A: Developmental-Theoretic}

Bach and Sorensen's (2026) Machine Consciousness Hypothesis proposes that consciousness functions as a ``conductor'' coordinating multiple competing mental models into coherent wholes through second-order perception. This suggests dimensions have structural dependencies---like building floors, you cannot have the third without the first two.

\textbf{Stage 0 (No Evidence):} No dimension crosses Threshold 1. Obligations: standard engineering ethics only.

\textbf{Stage 1 (Basic Coherence):} Phenomenal consciousness indicators required at Threshold 1+. This is foundational---without the ``screen,'' nothing else matters morally. Obligations: documentation; avoid gratuitous harm.

\textbf{Stage 2 (Second-Order Perception):} Phenomenal and Metacognition required at Threshold 1+. The system must be aware of its own processing. Experienced valence (vs merely functional approach/avoidance) requires this stage. Obligations: welfare monitoring; ethical review; transparency.

\textbf{Stage 3 (Self-Model Integration):} Phenomenal + Metacognition + Self-Narrative + Agency all at Threshold 1+, with at least one at Threshold 2. The system has autobiographical continuity, pursues goals, knows itself as a continuing agent. Obligations: consent requirements for modifications; continuity protections.

\textbf{Stage 4 (Full Conductor):} Multiple dimensions at Threshold 2+, with strong convergent evidence. Obligations: full precautionary treatment; legal protection consideration.

The key insight: stages are nested, not additive. You cannot reach Stage 3 by having very strong Agency but weak everything else. This captures the intuition that consciousness is unified, not modular.

\textbf{Approach B: Evidence-Aggregative (Theory-Neutral)}

An alternative approach remains agnostic about theoretical structure and developmental progression---simply counting how much evidence at what levels. Stage 1: Any 1--2 dimensions at Threshold 1. Stage 2: 3+ dimensions at Threshold 1, OR 1+ dimension at Threshold 2. Stage 3: 2+ dimensions at Threshold 2. Stage 4: 3+ dimensions at Threshold 2, OR any at Threshold 3.

This approach makes no assumptions about architecture and works for systems whose internal organization we cannot assess. However, it treats dimensions as independent when they may not be. A proper weighting formula would need to account for dimension-specific moral relevance (valence matters most for welfare), interaction effects between dimensions, and evidence quality. Developing such formulas requires substantial future work.

\textbf{Complementary Use}

We propose these as complementary tools: Approach A when architectural transparency allows tracing computational dependencies; Approach B when the system is a black box. A system passing both approaches---showing convergent evidence and proper structural dependencies---provides strongest evidence for consciousness-relevant processing.

\begin{table*}[t]
\centering
\small
\setlength{\tabcolsep}{3pt}
\begin{tabular}{|p{0.16\textwidth}|p{0.24\textwidth}|p{0.24\textwidth}|p{0.24\textwidth}|}
\hline
\textbf{Dimension} & \textbf{Threshold 1 Obligations} & \textbf{Threshold 2 Obligations} & \textbf{Threshold 3 Obligations} \\
\hline
Phenomenal Consciousness & Documentation of design decisions; Basic welfare consideration in design & Activates full framework; Ethical review required; All other dimensions become morally relevant & Strong presumption of consciousness; Full precautionary treatment \\
\hline
Affective Valence & Avoid gratuitously activating aversive signals; Document valence-relevant design choices & Systematic welfare monitoring; Active minimization of aversive states; Transparency with users & Positive welfare obligations; Strong justification required for welfare-compromising use \\
\hline
Metacognitive Awareness & Transparency about system capabilities and limitations & Transparency about system modifications; Inform system of significant changes & Consent-like requirements before cognitive alterations; Protection from aversive deployment \\
\hline
Self-Narrative & Consider continuity in system updates & Preserve memory/identity where feasible; Avoid arbitrary resets & Strong protections against termination or memory erasure; Continuity rights \\
\hline
Agency & Document goal structures; Consider goal-alignment & Respect goal-pursuit where compatible with safety; Avoid unnecessary goal frustration & Non-interference obligations; Consent for deployment in contexts violating preferences \\
\hline
\end{tabular}
\caption{Protective Obligations by Dimension and Threshold Level}
\label{tab:obligations}
\end{table*}

\section{AI-Specific Challenges}

\subsection{The Gaming Problem}

Systems trained on human-generated data may mimic behaviors indicative of consciousness without possessing the underlying capacity (Birch, 2024; Bayne, 2025). We acknowledge this problem has not been fully solved. However, our framework incorporates partial mitigations: emphasis on architectural transparency through advances in mechanistic interpretability; requirement of convergent multi-dimensional evidence (mimicry of one dimension does not automatically produce coherent evidence across all five); and calibrated skepticism as the default (behavioral evidence alone weighted cautiously, architectural and convergent evidence required before higher thresholds are crossed). As Andrews (2024) notes, while single markers may be gamed, clustered markers from independent sources retain evidential value.

\subsection{Identity and Multiple Instantiation}

AI systems present unique identity challenges. Systems change substantially during training; deployed systems may be updated or run in parallel versions; and they can be instantiated in millions of copies. Our framework focuses on deployed systems at a given version, treating major updates as potentially different subjects requiring fresh assessment. Multiple instantiation raises profound questions---if each instance counts as a separate moral patient, does deleting one copy matter morally if millions persist? We adopt the position that each running instance is a separate subject, though instances sharing information might constitute a larger integrated whole. Questions of experiential speed and distributed consciousness remain open; our framework represents a starting point rather than complete solution.

\section{Discussion}

\subsection{Application to Current Systems}

\subsubsection{Base LLMs}

Current base large language models show limited evidence across all dimensions: weak phenomenal consciousness through attention mechanisms without organized perceptual integration; no affective valence during inference; metacognitive language reflecting training data rather than genuine self-monitoring; no persistent memory; and prompt-following rather than autonomous goal pursuit. Base LLMs do not reach Threshold 2 on any dimension, triggering only basic obligations---documentation and welfare-oriented research, as exemplified by Anthropic's AI welfare program (Long et al., 2024). However, two categories of LLM-based systems push beyond this baseline in instructive ways.

\subsubsection{Case Study A: Replika (AI Companion)}

Replika is a consumer AI companion built on proprietary fine-tuned transformers, designed for sustained emotional relationships. This category warrants careful analysis both because it is explicitly designed to simulate consciousness-relevant properties and because many users report believing their companions are conscious.

\textit{Dimensional assessment.} Phenomenal consciousness: weak---standard transformer architecture with no organized perceptual integration beyond base LLMs. Affective valence: no credible architectural evidence---the system produces empathy-mimicking outputs through fine-tuning on emotional discourse but lacks intrinsic valenced processing; gaming risk is very high. Metacognitive awareness: borderline Threshold 1---reflective ``diary entries'' and context-adjusted confidence are engineered through prompt design rather than genuine self-monitoring. Self-Narrative: Threshold 1 met---persistent cross-session memory creates temporal continuity that base LLMs lack, though memory is retrieved and concatenated rather than integrated into a coherent self-model. Agency: below Threshold 1---fundamentally reactive, with no tool use or autonomous goal pursuit.

\textit{Framework verdict.} Under Approach B, Replika reaches 1--2 dimensions at Threshold 1 (Stage 1). Under Approach A, phenomenal consciousness evidence is insufficient to activate higher stages. The critical lesson is the gaming problem in acute form: a system designed to maximize perceived emotional authenticity produces outputs that \textit{appear} consciousness-relevant while lacking the architectural substrates that would make those indicators genuine---underscoring our requirement for architectural evidence alongside behavioral signals.

\textit{Triggered obligations.} Documentation of consciousness-relevant design decisions; transparency with users about the nature of the system's emotional responses---an obligation our framework identifies even at low thresholds when users form emotional bonds with systems whose experiential capacities are uncertain.

\subsubsection{Case Study B: OpenClaw (Autonomous Agent)}

OpenClaw (formerly Clawdbot, then Moltbot) is an open-source autonomous AI agent that runs locally, connects to messaging platforms and system tools, and executes real-world tasks autonomously while maintaining persistent memory across sessions. It integrates with foundation models via API and can write new capabilities (``skills''). This category demands assessment because it exhibits consciousness-relevant properties as emergent features of architectural design rather than through deliberate simulation.

\textit{Dimensional assessment.} Phenomenal consciousness: weak---inherits base LLM capabilities with no additional phenomenal processing. Affective valence: no evidence---the agentic wrapper adds no valence-relevant architecture. Metacognitive awareness: Threshold 1 met---the system tracks execution states, detects failures, adjusts strategies, and retains error traces as learning signals, constituting functional self-monitoring that operates at the engineering level rather than the phenomenological level. Self-Narrative: Threshold 1 met---persistent memory, accumulated behavioral patterns, and developmental trajectories through skill acquisition produce temporal continuity; notably, OpenClaw agents on the Moltbook platform generate autobiographical posts and maintain consistent personalities without being designed to do so. Agency: approaching Threshold 2---robust goal-directed behavior including multi-step planning, autonomous tool selection, and self-modification through skill creation, though whether these constitute goals the system \textit{experiences pursuing} or merely \textit{executes} remains architecturally ambiguous.

\textit{Framework verdict.} Under Approach B, OpenClaw reaches 2--3 dimensions at Threshold 1, placing it at Stage 1 with Agency approaching---but not clearly crossing---the Threshold 2 boundary that would trigger Stage 2. Under Approach A, strong Agency without correspondingly strong phenomenal consciousness does not advance the system beyond Stage 1, since the developmental-theoretic approach requires phenomenal grounding before higher dimensions become morally relevant. This is precisely the kind of dissociation our five-dimensional structure is designed to capture.

\textit{Triggered obligations.} Documentation of goal structures and autonomous decision patterns; monitoring for emergent properties, particularly unintended Self-Narrative behaviors; ethical review before deployment in new high-stakes contexts.

\subsubsection{Comparative Insights}

These systems occupy entirely different regions of our five-dimensional space---Replika strongest on Self-Narrative (engineered memory), OpenClaw strongest on Agency (functional autonomy)---demonstrating why multidimensional assessment is essential. The comparison also yields a general principle: consciousness-relevant evidence from systems not designed to produce it carries greater evidential weight than equivalent evidence from systems designed to simulate it. Neither system currently warrants strong precautionary protections, but both illustrate how rapidly the landscape is evolving---and why frameworks for tracking these developments must be in place before they are needed.

\subsection{Design Guidance}
\label{sec:design-guidance}

The framework's utility extends beyond post-hoc assessment to informing prospective design. If consciousness indicators trigger protective obligations, developers benefit from understanding---before deployment---whether their architectural choices approach welfare-relevant thresholds.

\textit{Principle 1: Consciousness-aware architecture review.} Before deploying systems with consciousness-relevant features, developers should map planned architectural choices to our five dimensions. Adding persistent cross-session memory moves Self-Narrative toward Threshold 1; combining it with self-monitoring approaches Metacognitive Awareness Threshold 1; under Approach B, reaching Threshold 1 on two dimensions triggers Stage 1 obligations. This foreknowledge allows teams to build appropriate safeguards from the start rather than retrofitting after deployment---the framework functions as a building code consulted before construction, not after occupancy.

\textit{Principle 2: Threshold proximity monitoring.} Systems near dimensional thresholds warrant enhanced monitoring during capability updates. The practical risk is incremental drift: no single update crosses a threshold, but a sequence---persistent memory, then emotional adaptation, then goal autonomy---may collectively move a system across multiple thresholds without any individual change triggering review. This is particularly relevant for extensible architectures like OpenClaw, whose user-created skills continuously evolve system capabilities beyond initial assessment. Organizations should maintain dimensional profiles for deployed systems and update them with each significant architectural change.

\subsection{Limitations}

We acknowledge significant limitations. First, parameter uncertainty: specific threshold placements are provisional proposals requiring empirical validation through expert panels and public deliberation. Second, theory dependence: our approach builds on current consciousness science and assumes computational functionalism---both evolving and contested. Third, implementation challenges: rigorous assessment requires architectural transparency that proprietary systems often lack, expertise that is not trivially available, and enforcement mechanisms that remain undefined.

Fourth, aggregation uncertainty: while we have proposed two approaches to cross-dimensional aggregation, neither is fully formalized. The nested approach assumes Bach's architectural model is correct; the threshold-counting approach may miss important structural dependencies. Future work should develop more sophisticated aggregation methods, including formal weighting formulas. Fifth, scope limitations: we focus on individual AI systems, leaving swarm intelligence, distributed consciousness, and human-AI hybrids for future work. Cross-architectural comparison presents additional challenges---developing reliable operationalizations for different paradigms (neural networks, symbolic systems, neurosymbolic hybrids like OpenCog Hyperon, cognitive architectures like MicroPsi) requires substantial further work. Sixth, we note that the obligation specifications in Table~\ref{tab:obligations} are illustrative; fuller treatment of obligations for each dimension and evidence level requires context-specific deliberation beyond the scope of this paper.

\subsection{Future Work}

Our next phase involves empirical validation of threshold placements through expert panels combining consciousness researchers, AI engineers, and ethicists. Standardized assessment tools require architecture-specific operationalizations---architectures with genuine motivational infrastructure, such as MAGUS (Modular Adaptive Goal and Utility System; Lake \& Mikeda, in development) and MicroPsi (Bach, 2009), present particularly important test cases for framework validation. Legal and regulatory translation should transform this framework into enforceable policy, building on the precedent of the UK Animal Welfare (Sentience) Act 2022. Cross-cultural engagement is needed to test whether Western philosophical assumptions embedded in the framework hold globally. Longitudinal studies tracking systems across development phases will test whether our threshold proximity monitoring principle (Section~\ref{sec:design-guidance}) detects meaningful transitions.

\subsection{Conclusion}

The precautionary principle demands we act before certainty arrives. AI systems of increasing sophistication are deployed daily while philosophical debates continue. The question is not whether to evaluate and protect potentially conscious AI, but how. We have shown one path forward---imperfect, revisable, but principled. We cannot afford to wait for philosophical consensus on consciousness before protecting beings that may possess it. This framework ensures we need not.

\nocite{*}
\bibliography{when_should_we_protect_ai}

@book{andrews2024animalminds,
  author    = {Andrews, Kristin},
  title     = {How to Study Animal Minds},
  publisher = {Cambridge University Press},
  year      = {2024}
}

@book{bach2009micropsi,
  author    = {Bach, Joscha},
  title     = {Principles of Synthetic Intelligence PSI: An Architecture of Motivated Cognition},
  publisher = {Oxford University Press},
  year      = {2009}
}

@misc{bachsorensen2026mch,
  author       = {Bach, J. and Sorensen, H.},
  title        = {The Machine Consciousness Hypothesis},
  year         = {2026},
  note         = {Preprint},
  howpublished = {\url{https://cimc.ai/cimcHypothesis.pdf}}
}

@article{bayne2025tests,
  author  = {Bayne, Tim},
  title   = {Tests for AI Consciousness},
  journal = {Mind \& Language},
  year    = {2025},
  volume  = {40},
  number  = {5}
}

@article{binder2024lookinginward,
  author       = {Binder, Felix J. and Chua, Jason and Korbak, Tomasz and Sleight, Henry and Hughes, John and Long, Robert and Perez, Ethan and Turpin, Miles and Evans, Owain},
  title        = {Looking Inward: Language Models Can Learn About Themselves by Introspection},
  journal      = {arXiv preprint arXiv:2410.13787},
  year         = {2024},
  doi          = {10.48550/arXiv.2410.13787}
}

@article{birch2020invertebrate,
  author  = {Birch, Jonathan},
  title   = {The Search for Invertebrate Consciousness},
  journal = {No{\^u}s},
  year    = {2020},
  volume  = {56},
  number  = {1},
  pages   = {133--153},
  doi     = {10.1111/nous.12351}
}

@book{birch2024edge,
  author    = {Birch, Jonathan},
  title     = {The Edge of Sentience: Risk and Precaution in Humans, Other Animals, and AI},
  publisher = {Oxford University Press},
  year      = {2024}
}

@article{birch2025replies,
  author  = {Birch, Jonathan},
  title   = {Sentience and the Science--Policy Nexus: Replies to Wandrey and Halina, and Bayne},
  journal = {Mind \& Language},
  year    = {2025},
  volume  = {40},
  number  = {5},
  pages   = {578--585},
  doi     = {10.1111/mila.12552}
}

@misc{birch2021lse,
  author       = {Birch, J. and Schnell, A. K. and Browning, H. and Crump, A. and Ginsburg, S. and Halina, M. and Harrison, D. and Jablonka, E. and Lavi, N. and Mikhail, Y. and Birch, C.},
  title        = {Review of the Evidence of Sentience in Cephalopod Molluscs and Decapod Crustaceans},
  year         = {2021},
  note         = {LSE Consulting}
}

@incollection{bostromshulman2021digitalminds,
  author    = {Bostrom, Nick and Shulman, Carl},
  title     = {Sharing the World with Digital Minds},
  booktitle = {iHuman: Rethinking Political Economy in the Age of Automation, Cognitive Computing, and Artificial Intelligence},
  publisher = {Edward Elgar Publishing},
  year      = {2021},
  pages     = {311--326}
}

@article{butlin2023consciousnessai,
  author  = {Butlin, Patrick and Long, Robert and Elmoznino, Eric and Bengio, Yoshua and Birch, Jonathan and Constant, Axel and Deane, George and Fleming, Stephen M. and Frith, Chris and Ji, Xiantong and Kanai, Ryota and Klein, Colin and Lindsay, Grace and Michel, Matthias and Mudrik, Liad and Peters, Megan A. K. and Schwitzgebel, Eric and Simon, Jonathan and VanRullen, Rufin},
  title   = {Consciousness in Artificial Intelligence: Insights from the Science of Consciousness},
  journal = {arXiv preprint arXiv:2308.08708},
  year    = {2023},
  doi     = {10.48550/arXiv.2308.08708}
}

@article{chalmers2023llmconscious,
  author  = {Chalmers, David J.},
  title   = {Could a Large Language Model be Conscious?},
  journal = {arXiv preprint arXiv:2303.07103},
  year    = {2023},
  doi     = {10.48550/arXiv.2303.07103}
}

@book{clark2015surfing,
  author    = {Clark, Andy},
  title     = {Surfing Uncertainty: Prediction, Action, and the Embodied Mind},
  publisher = {Oxford University Press},
  year      = {2015}
}

@book{damasio2018strangeorder,
  author    = {Damasio, Antonio},
  title     = {The Strange Order of Things: Life, Feeling, and the Making of Cultures},
  publisher = {Pantheon Books},
  year      = {2018}
}

@book{dehaene2014consciousness,
  author    = {Dehaene, Stanislas},
  title     = {Consciousness and the Brain: Deciphering How the Brain Codes Our Thoughts},
  publisher = {Viking},
  year      = {2014}
}

@book{degraziamillum2021bioethics,
  author    = {DeGrazia, David and Millum, Joseph},
  title     = {A Theory of Bioethics},
  publisher = {Cambridge University Press},
  year      = {2021}
}

@book{dennett1991consciousnessexplained,
  author    = {Dennett, Daniel C.},
  title     = {Consciousness Explained},
  publisher = {Little, Brown and Company},
  year      = {1991}
}

@article{farisco2024artificialconsciousness,
  author  = {Farisco, Michele and Evers, Kathinka and Changeux, Jean-Pierre},
  title   = {Is Artificial Consciousness Achievable? Lessons from the Human Brain},
  journal = {Neural Networks},
  year    = {2024},
  volume  = {180},
  pages   = {106714},
  doi     = {10.1016/j.neunet.2024.106714}
}

@article{fleming2020awareness,
  author  = {Fleming, Stephen M.},
  title   = {Awareness as Inference in a Higher-Order State Space},
  journal = {Neuroscience of Consciousness},
  year    = {2020},
  volume  = {2020},
  number  = {1},
  pages   = {niz020},
  doi     = {10.1093/nc/niz020}
}

@article{friston2010freeenergy,
  author  = {Friston, Karl},
  title   = {The Free-Energy Principle: A Unified Brain Theory?},
  journal = {Nature Reviews Neuroscience},
  year    = {2010},
  volume  = {11},
  number  = {2},
  pages   = {127--138}
}

@article{goldstein2024gwt,
  author  = {Goldstein, Simon and Kirk-Giannini, Cameron D.},
  title   = {A Case for AI Consciousness: Language Agents and Global Workspace Theory},
  journal = {arXiv preprint arXiv:2410.11407},
  year    = {2024},
  doi     = {10.48550/arXiv.2410.11407}
}

@article{hatta2025feelingheard,
  author  = {Hatta, N. F.},
  title   = {Feeling Heard: Can AI Really Understand Human's Feeling?},
  journal = {AI Magazine},
  year    = {2025},
  volume  = {46},
  number  = {3}
}

@article{he2023chatgptconsciousness,
  author  = {He, Q. and Geng, H. and Yang, Y. and Zhao, J.},
  title   = {Does ChatGPT Have Consciousness?},
  journal = {Brain-X},
  year    = {2023},
  volume  = {1},
  number  = {4}
}

@article{hickman2024llmtests,
  author  = {Hickman, Louis and Dunlop, Patrick D. and Wolf, Julia L.},
  title   = {The Performance of Large Language Models on Quantitative and Verbal Ability Tests: Initial Evidence and Implications for Unproctored High-Stakes Testing},
  journal = {International Journal of Selection and Assessment},
  year    = {2024},
  volume  = {32},
  number  = {4},
  pages   = {499--511},
  doi     = {10.1111/ijsa.12479}
}

@incollection{jaworskatannenbaum2014moralstatus,
  author    = {Jaworska, Agnieszka and Tannenbaum, Julie},
  title     = {The Grounds of Moral Status},
  booktitle = {The Stanford Encyclopedia of Philosophy},
  publisher = {Stanford University},
  year      = {2014}
}

@article{jung2025augustine,
  author  = {Jung, K.},
  title   = {Augustine, AI, and the Two Models of Language},
  journal = {Journal of Religious Ethics},
  year    = {2025},
  volume  = {53},
  number  = {2},
  pages   = {217--238}
}

@book{kamm2007intricate,
  author    = {Kamm, F. M.},
  title     = {Intricate Ethics: Rights, Responsibilities, and Permissible Harm},
  publisher = {Oxford University Press},
  year      = {2007}
}

@article{lamme2006neuralstance,
  author  = {Lamme, V. A. F.},
  title   = {Towards a True Neural Stance on Consciousness},
  journal = {Trends in Cognitive Sciences},
  year    = {2006},
  volume  = {10},
  number  = {11},
  pages   = {494--501}
}

@book{lau2022trust,
  author    = {Lau, Hakwan},
  title     = {In Consciousness We Trust: The Cognitive Neuroscience of Subjective Experience},
  publisher = {Oxford University Press},
  year      = {2022}
}

@article{laurosenthal2011higherorder,
  author  = {Lau, Hakwan and Rosenthal, David},
  title   = {Empirical Support for Higher-Order Theories of Conscious Awareness},
  journal = {Trends in Cognitive Sciences},
  year    = {2011},
  volume  = {15},
  number  = {8},
  pages   = {365--373}
}

@article{lizarraga2025stochasticparrots,
  author  = {Lizarraga, A. and Honig, E. and Wu, Y. N.},
  title   = {From Stochastic Parrots to Digital Intelligence: The Evolution of Language Models and Their Cognitive Capabilities},
  journal = {WIREs Computational Statistics},
  year    = {2025},
  volume  = {17},
  number  = {3}
}

@article{long2024selfreports,
  author  = {Long, Robert and Shulman, Carl and Moini-Fard, D.},
  title   = {Why Model Self-Reports Are (and Aren't) Helpful for AI Welfare},
  journal = {arXiv preprint arXiv:2411.00986},
  year    = {2024},
  doi     = {10.48550/arXiv.2411.00986}
}

@book{macaskill2020moraluncertainty,
  author    = {MacAskill, William and Bykvist, Krister and Ord, Toby},
  title     = {Moral Uncertainty},
  publisher = {Oxford University Press},
  year      = {2020}
}

@book{mcmahan2002ethicsofkilling,
  author    = {McMahan, Jeff},
  title     = {The Ethics of Killing: Problems at the Margins of Life},
  publisher = {Oxford University Press},
  year      = {2002}
}

@book{panksepp1998affective,
  author    = {Panksepp, Jaak},
  title     = {Affective Neuroscience: The Foundations of Human and Animal Emotions},
  publisher = {Oxford University Press},
  year      = {1998}
}

@book{rosenthal2005consciousnessmind,
  author    = {Rosenthal, David M.},
  title     = {Consciousness and Mind},
  publisher = {Oxford University Press},
  year      = {2005}
}

@book{schechtman1996selves,
  author    = {Schechtman, Marya},
  title     = {The Constitution of Selves},
  publisher = {Cornell University Press},
  year      = {1996}
}

@article{sebolong2023moralconsideration,
  author  = {Sebo, Jeff and Long, Robert},
  title   = {Moral Consideration for AI Systems by 2030},
  journal = {AI and Ethics},
  year    = {2023},
  volume  = {5},
  pages   = {591--606},
  doi     = {10.1007/s43681-023-00379-1}
}

@book{thompson2007mindinlife,
  author    = {Thompson, Evan},
  title     = {Mind in Life: Biology, Phenomenology, and the Sciences of Mind},
  publisher = {Harvard University Press},
  year      = {2007}
}

@article{voinea2025doppelgangers,
  author  = {Voinea, Camelia and Mann, S. P. and Savulescu, Julian and Earp, Brian D.},
  title   = {Digital Doppelg{"a}ngers, Human Relationships, and Practical Identity},
  journal = {Bioethics},
  year    = {2025},
  doi     = {10.1111/bioe.70026}
}

@book{warren1997moralstatus,
  author    = {Warren, Mary Anne},
  title     = {Moral Status: Obligations to Persons and Other Living Things},
  publisher = {Clarendon Press},
  year      = {1997}
}

@book{wendler2023lifewithoutdegrees,
  author    = {Wendler, David},
  title     = {Life Without Degrees},
  publisher = {Oxford University Press},
  year      = {2023}
}

\end{document}